\documentclass[sigconf,10pt]{acmart} 
\renewcommand\footnotetextcopyrightpermission[1]{} 
\AtBeginDocument{%
  \providecommand\BibTeX{{%
    \normalfont B\kern-0.5em{\scshape i\kern-0.25em b}\kern-0.8em\TeX}}}

\usepackage[inline]{enumitem}

\newcommand{\fakeparagraph}[1]{\vspace{.5em}\noindent\textbf{#1}.\hspace{0.25em}}
\newcommand{\sysname}{FSL-GAN }

\begin{document}

\title{Federated Split GANs}

\author{Pranvera Korto\c{c}i}
\orcid{0000-0003-4006-8969}
\affiliation{%
  \institution{University of Helsinki}
  \city{Helsinki}
  \country{Finland}
}
\email{pranvera.kortoci@helsinki.fi}

\author{Yilei Liang}
\affiliation{%
  \institution{University of Cambridge}
  \city{Cambridge}
  \country{UK}}
\email{yl841@cst.cam.ac.uk}

\author{Pengyuan Zhou}
\affiliation{%
  \institution{USTC}
  \city{Hefei}
  \country{China}
}
\email{pyzhou@ustc.edu.cn}

\author{Lik-Hang Lee}
\affiliation{%
 \institution{KAIST}
 \city{Daejeon}
 \country{South Korea}}
\email{likhang.lee@kaist.ac.kr}

\author{Abbas Mehrabi}
\affiliation{%
  \institution{Northumbria University}
  \city{Newcastle}
  \country{UK}}
\email{abbas.mehrabidavoodabadi}
 \email{@northumbria.ac.uk}

\author{Pan Hui}
\affiliation{%
  \institution{HKUST}
  \city{Hong Kong SAR}
  \country{China}
  }
\email{panhui@cse.ust.hk}

\author{Sasu Tarkoma}
\affiliation{%
  \institution{University of Helsinki}
  \city{Helsinki}
  \country{Finland}
  }
\email{sasu.tarkoma@helsinki.fi}

\author{Jon Crowcroft}
\affiliation{%
  \institution{University of Cambridge}
  \city{Cambridge}
  \country{UK}
  }
\email{jon.crowcroft@cl.cam.ac.uk}

\renewcommand{\shortauthors}{Korto\c{c}i, et al.}

\begin{abstract}
Mobile devices and the immense amount and variety of data they generate are key enablers of machine learning (ML)-based applications.
Traditional ML techniques have shifted toward new paradigms such as federated (FL) and split learning (SL) to improve the protection of user’s data privacy. However, these paradigms often rely on server(s) located in the edge or cloud to train computationally-heavy parts of a ML model to avoid draining the limited resource on client devices, resulting in exposing device data to such third parties. This work proposes an alternative approach to train computationally-heavy ML models in user’s devices themselves, where corresponding device data resides. Specifically, we focus on GANs (generative adversarial networks) and leverage their inherent privacy-preserving attribute. We train the discriminative part of a GAN with raw data on user’s devices, whereas the generative model is trained remotely (e.g., server) for which there is no need to access sensor true data. Moreover, our approach ensures that the computational load of training the discriminative model is shared among user’s devices -- proportional to their computation capabilities -- by means of SL. We implement our proposed collaborative training scheme of a computationally-heavy GAN model in real resource-constrained devices. The results show that our system preserves data privacy, keeps a short training time, and yields same accuracy of model training in a unconstrained devices (e.g., cloud). Our code can be found on https://github.com/YukariSonz/FSL-GAN
\end{abstract}

\keywords{federated learning, GAN, split learning}

\maketitle

\section{Introduction}
\label{sec:intro}

The use of mobile and ubiquitous devices such as smartphones and other handheld  devices equipped with a wide range of sensors has given rise to a massive amount of collected data. The data consists of both sensor data, which relates to phenomena around us, as well as data related to the individuals carrying such sensors. Many applications -- such as those in the field of smart cities, smart transportation, manufacturing, supply chain, healthcare, and security -- are built upon such data. The vast proliferation of application domains results in a huge amount of data being collected. Traditionally such data is stored in the cloud for further processing and analysis. However, such centralized data analytics in the cloud, for instance, poses several data privacy issues~\cite{singh2017cloud}. Specifically, the cloud would have access to users' privacy-sensitive data, which could potentially disclose their daily paths, home location, or even financial-related information. As such, fully-centralized data collection and analysis does not provide needed data privacy and security. By contrast to such traditional AI techniques, distributed collaborative learning solutions such as federated learning (FL) and split learning (SL) provide data privacy~\cite{turina2020combining}. Moreover, SL is particularly suitable for training computationally-heavy learning models on resource-constrained devices. As such, FL and SL combined~\cite{kairouz2019advances} enable a whole range of data-sensitive applications in medicine, finance, and extended reality (XR) on resource-constrained devices.

XR applications are now operating in different domains such as industry 4.0~\cite{gattullo2019towards}, smart transportation~\cite{zhou2020edge}, education~\cite{doolani2020review}, and needless to say, gaming, to create immersive and interactive experiences. Most of these applications have stringent delay requirements and collect sensitive data related to user privacy, such as real-time eye tracking and context aware sensing. As such, keeping the processing of data at local devices have great potential for enabling these applications. However, XR applications are notably computationally-heavy, relying mostly on deep neural networks to learn data models, which restricts their applicability in mobile devices due to their constrained resources.


In this article we envision a system that combines both FL and SL to enable training of generative adversarial networks (GANs)~\cite{goodfellow2020generative} with resource-constrained devices. A GAN comprises a generative model $G$ that captures data distribution and a discriminative model $D$ that judges if the sample data came from the raw user data or was generated by $G$. GANs outperform traditional ML algorithms in feature learning and representation~\cite{cao2018recent} and have been widely adopted for computer vision tasks which are at the core of XR applications. However, GAN algorithms are computationally-heavy, which greatly hinders their training with resource-constrained devices. Moreover, GAN can potentially raise the privacy leakage risk~\cite{liu2019ppgan}. To address such issues, we use FL to ensure data privacy by having clients train a learning model on their own data locally, without sharing such data among clients and a remote server. In addition, we complement FL with SL to distribute the training workload among user devices. 
As such, each device trains only a subset of layers of the \emph{split} network model~\cite{thapa2020splitfed}.


Our work builds on the premise that there is an ever-growing proliferation of XR applications and services, which will become even more relevant due to technological advancements on 5G and beyond networks. As such, we need to provide dependable and resource-efficient methods to train computationally-heavy models on resource-constrained devices, with a focus on data privacy. Specifically, our contributions in this work are threefold as follows.
\begin{enumerate}
    \item We apply FL and SL to the discriminator of a GAN to split a training model into portions, and assign them across multiple devices per client (if available) to achieve more granular task assignment than related works.
    \item We devise a heuristic device selection method that turns device heterogeneity into an advantage by assigning heavier model portions to devices whose computational capabilities support it.
    \item We run simulations on a controlled environment by emulating resource-constrained devices and their computational capabilities. The results show that introducing \emph{device capability-awareness} along with granular task assignment reduces the training time of a discriminator, while achieving a high model accuracy.
\end{enumerate}

The rest of the paper is structured as follows. Section~\ref{sec:rel_work} discusses related works and emphasizes the motivation of our work. Section~\ref{sec:background} introduces our system model and formulates the problem, whereas Section~\ref{sec:fsl-gan} details the architecture of our proposed FSL-GAN solution. Section~\ref{sec:evaluation} evaluates the performance of our FL and SL method applied to GANs. Finally, Section~\ref{sec:conclusions} concludes the work with some final remarks.

\section{Related Works}
\label{sec:rel_work}

GANs~\cite{goodfellow2020generative} are a subset of machine learning models used in semi-supervised and unsupervised learning. GANs are successfully deployed in artificial intelligence (AI) applications such as medicine, computer vision, vocal AI, autonomous driving, and natural language processing~\cite{lin2017adversarial,hsu2017voice,mogren2016c,jabbar2021survey}. Both the \emph{generator} $G$ and the \emph{discriminator} $D$ of a GAN are deep neural networks; in particular, $D$ is a classifier that aims at differentiating between real data stored on-premises (on the device) and fake data generated by $G$. As such, it is highly desirable, and in some situations necessary, to train $D$ on the devices that generate and host the real (true) data. However, such devices come with different computational capabilities which might hinder the on-site training of GAN models. Enabling distributed learning of GAN models, especially for XR applications, requires to tackle two main challenges: 
\begin{enumerate*}[label=(\roman*)]
  \item constrained resources (i.e., heterogeneity) on the devices where we train the model, and
  \item data privacy.
\end{enumerate*}

Medical applications, for instance, are built upon highly privacy-sensitive patient data, and thus any ML model run on medical data must ensure its privacy. The work in~\cite{chang2020multi} proposes a distributed asynchronized discriminator GAN (AsynDGAN) framework where each data source (e.g., device) trains its discriminator $D$ with its own local data. As such, it provides data privacy and security due to data isolation among devices and the generator. Similarly,~\cite{rasouli2020fedgan} proposes FedGAN, a distributed framework to train a GAN across different data sources and corresponding devices. FedGAN trains both the $D$ and $G$ locally, while an intermediary entity averages the $D$ and $G$ parameters and broadcasts them back to the devices. While this approach ensures communication efficiency and data privacy, it unnecessarily trains $G$ on the devices, exhausting their resources. 

As the same time, FL and SL come to the rescue. FL provides data privacy by having each client train an ML model locally, with no data sharing among participating clients. However, constrained resources of participating clients greatly hinder the applicability of FL~\cite{leftout2022} -- clients might not be able to train an entire model. By contrast, SL addresses the limitations of FL by splitting the network architecture among participating clients and, possibly, a coordinating entity such as a server. For instance,~\cite{jeon2020privacy} proposes a privacy-sensitive parallel SL method where each client trains a portion of a model with a predefined size of mini batch that reflects the size of its (local) data set. Similarly,~\cite{abedi2020fedsl} proposes a combination scheme of FL and SL to train recurrent neural networks on distributed sequential data. This way, not only the clients have access to their own sequence of sequential data, ensuring data data privacy, but they also share the burden of training a model by splitting it, and thus ensuring model privacy. 

Our work, similar to existing research, successfully deploys FL and SL to ensure both data and model privacy. In addition, our proposed method optimally splits the network architecture of a model into smaller portions to be trained by individual devices, in proportion to their on-board computation capabilities. As such, our method assures that a model is trained within a given time budget by splitting a model in accordance with the computational budget of single devices.

\section{Background}
\label{sec:background}

\subsection{System Architecture}
\label{ssec:sys_architecture}

\begin{figure}[!t]
    \centering
    \includegraphics[width=.9\linewidth]{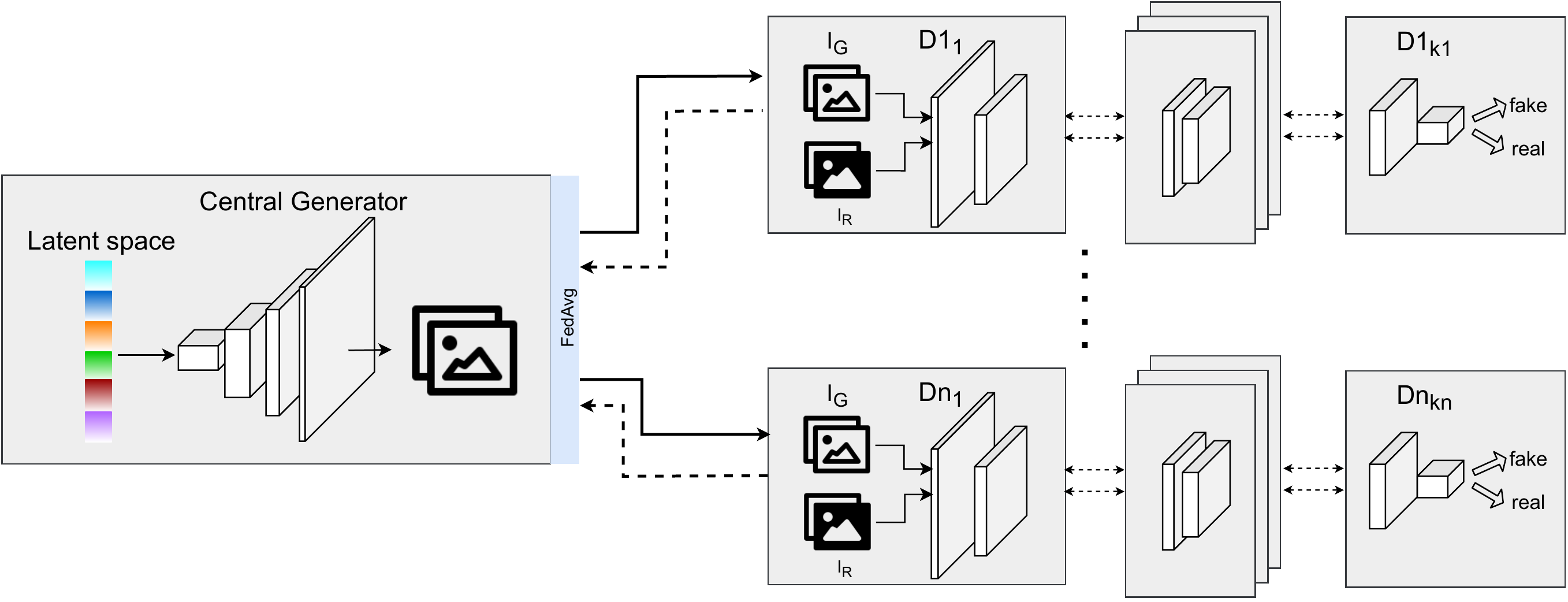}
    \caption{System architecture of our \sysname solution.}
    \label{fig:architecture}
\end{figure}

Our proposed distributed method of training a GAN via FL and SL is shown in Figure~\ref{fig:architecture}. The generator model is trained by a server that is usually located in the cloud. We assume that there are no constraints in terms of computational capabilities of the server. Moreover, we relief the computational burden on the resource-constrained devices by \emph{outsourcing} the training of the generator to a third party, while we still maintain data security. In fact, the generator generates fake data, which are not privacy-sensitive. By contrast, each client trains a discriminator network with their own local data, which is not shared among them or other entities. This approach resembles that of FL. Moreover, we assume that a client can have (i.e., own) more than one device. As such, the discriminator can be trained collaboratively among such devices, where each device trains only a split (portion) of the GAN, proportional to their on-board capabilities. This approach resembles the architecture of SL. We refer to our system architecture FSL-GAN, denoting a combined use of FL and SL approaches to train GANs.

Figure~\ref{fig:architecture} shows how a \emph{centralized} generator shares the generated images with the local discriminators, who in turn discriminate (compare) such images to their real, true data (images) and classify them as either fake or real. The parameters of the local discriminator models are then averaged in the same manner as FedAVG~\cite{mcmahan2017communication}.

\subsection{System Model}
\label{ssec:sys_model}

The system includes the set $\mathcal{C}=\{{c_1}, {c_2}, \dots, {c_M}\}$ of $M=|\mathcal{C}|$ FL clients, where each $c_m$ client has a set $\mathcal{D}=\{{d_1^m}, {d_2^m}, \dots, {d_N^m}\}$ of $N=|\mathcal{D}|$ SL devices, and a central server.
The generator model is trained by a server that is usually located in the cloud. We assume the \emph{central} generator shares the generated images with the local discriminators, who in turn discriminate (compare) such images to their real, true data and classify them as either fake or real. In our architecture, each client trains a discriminator network with their own data. The client's data resides on the client's devices only, and thus is not shared among different clients or other entities. As such, the architecture of each discriminator network is similar to that of FL. 

Training all the layers of the discriminator network model in one single device might not be feasible due to its limited resources in terms of memory, computational capabilities, and energy. As such, we opt to train a discriminator model among different devices of a given client. That is, computationally-heavy networks such as GANs are split among the devices of a client, where each device trains a portion, i.e., subset of layers, of the network. The split is done in accordance with the heterogeneous capabilities of single devices. Specifically, we consider two parameters, namely \textit{Time\_Factor} and \textit{Client\_Capacity}, to factor in the computation capability of the devices, and thus the time it takes to train a unit of network model, as well as the memory available on the devices.

\section{Device Selection in \sysname}
\label{sec:fsl-gan}

A successful deployment of \sysname relies on the condition that a client can train a GAN among their devices. As such, we take into account the device capabilities when splitting the model and assigning the corresponding portions to them. 

Our client selection follows two different approaches, as follows.
\begin{itemize}
  \item \textbf{Random selection.} 
  A client picks a device from their pool of available devices at \emph{random} and assigns a model training task (e.g., portion of the model) to it. A client can choose to either assign a
  \begin{itemize}
    \item \emph{single} portion of the model to the device, and continue by randomly selecting another device from the pool until the entire model is trained; or assign
    \item \emph{multiple} portions of the model to the same device as long as it can train them and store the corresponding results in the local memory. Similarly, the process continues until the entire model is trained.
  \end{itemize}
  A device is removed from the list of available devices if it cannot train any portion. Moreover, if a client does not have sufficient devices to train the entire model of a discriminator, it is removed from the list of clients that participate in FL.
  \item \textbf{Sort\_By\_Time(efficiency) selection.}  We incorporate the notions of on-board memory and local processing capabilities of a device -- \textit{Client\_Capacity} and \textit{Time\_Factor}, accordingly -- into a parameter called \emph{efficiency}. Such a parameter expresses the capacity of a device to train a model. We then sort the devices in decreasing order of \emph{efficiency}, and start assigning them portions of the network to train. Similar to the method mentioned earlier, we distinguish between assigning
  \begin{itemize}
    \item \emph{single}, or
    \item \emph{multiple} portions of the model to a single device.
  \end{itemize}
\end{itemize}

In addition, we compare our implementation of \sysname to a baseline scenario in which there is only one discriminator in the system. Contrary to a Deep Convolutional Generative Adversarial Network (DCGAN)~\cite{DBLP:journals/corr/RadfordMC15}, which also consists of a single discriminator, we consider a baseline scenario where the discriminator splits the model and assigns portions of such a model to their devices according to the \emph{Sort\_By\_Time(efficiency)} selection method.

\section{Evaluation}
\label{sec:evaluation}

We evaluate the performance of our proposed \sysname by measuring the time required for each epoch under different user selection methods (see Section~\ref{sec:fsl-gan}), and the benefits of using multiple discriminators at the same time. We compare the loss of the generator of our proposed scheme with that of a baseline scenario.

\fakeparagraph{Experimental Setup} 
We deployed our code on Google Colab Pro\footnote{\url{https://colab.research.google.com/}} equipped with Nvidia Tesla P100-PCIE-16GB GPU, and run all the models sequentially on the same GPU. To simulate the scenario of FL and SL in a heterogeneous environment, we set a parameter called \textit{Time\_Factor} to simulate devices with different training speeds, and \textit{Client\_Capacity} to simulate whether the device has enough resources (e.g., memory) to store the model. We use DCGAN~\cite{DBLP:journals/corr/RadfordMC15} with $3$ convolution layer blocks and the MNIST data set\footnote{\url{http://yann.lecun.com/exdb/mnist/}}. At each epoch, each client will sample $24$ batches of images (with BATCH\_SIZE = $256$) from the data set, i.e., $6144$ images for each client for one epoch. For the time benchmark, we enumerate $5$ clients with $4$ devices each with different capacities and processing power. We measure the computation time on the \textbf{slowest client}, which represents the bottleneck of the whole system. Note that at this point we ignore the communication overhead from clients to central generator, and only focus on the LAN communication overhead by different splitting strategies. For simplicity, we model the LAN communication time to $50$ ms for each LAN communication among devices of a single client. 

For the accuracy benchmark, we measure the generator loss for $500$ epochs by using a different number of discriminators, and the quality of images generated by the generator. By our evaluation we want to answer the following questions: 
\begin{enumerate*}[label=(\roman*)]
    \item what is the trade-off between time and device capacity under different model splitting strategies? and
    \item can the generator learn under multiple discriminator environments?
\end{enumerate*}


\fakeparagraph{Time Benchmark} 
\begin{figure}[!t]
    \centering
    \includegraphics[width=\linewidth]{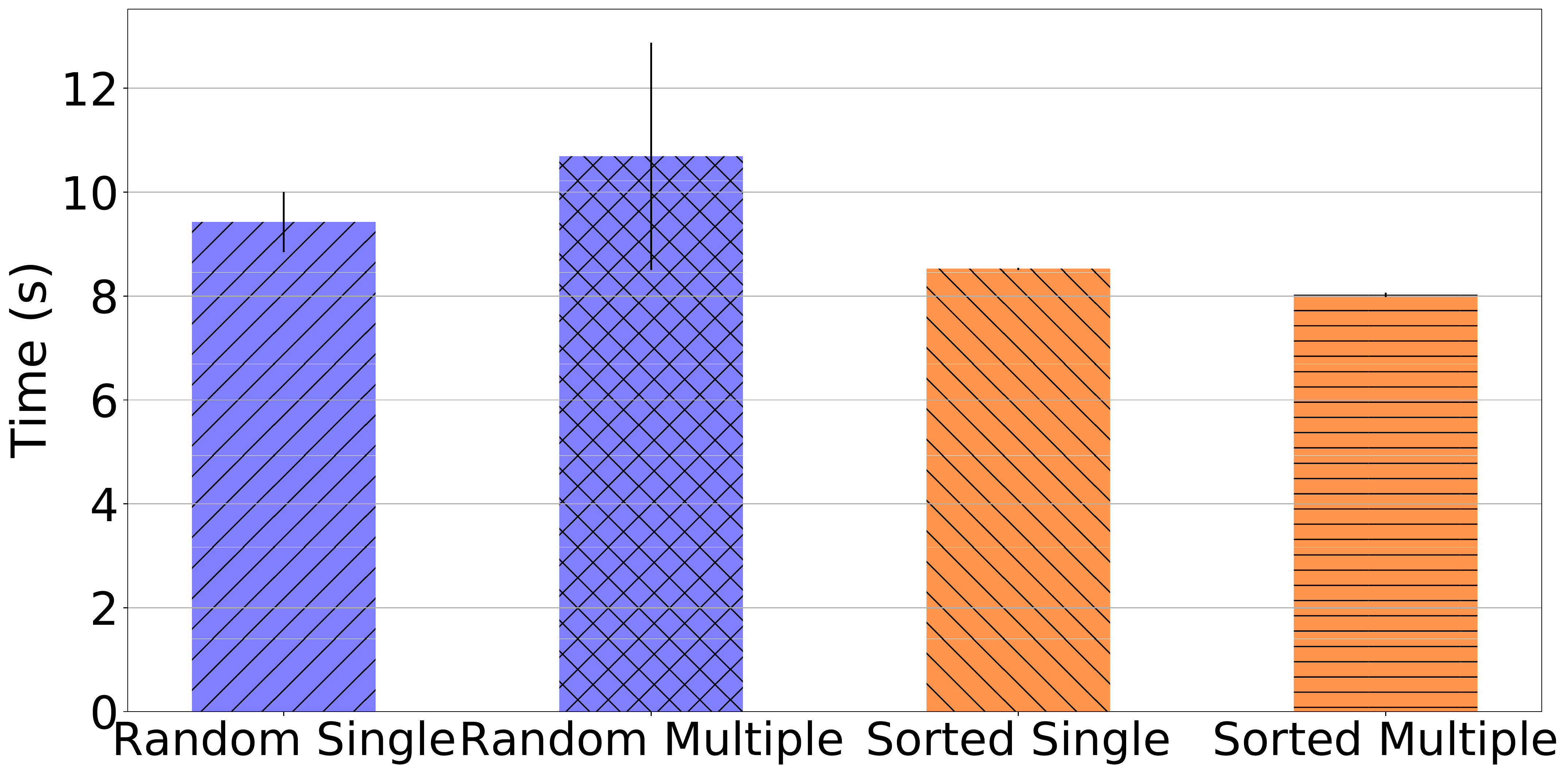}
    \caption{Time expenditure of the slowest discriminator in the network averaged over all epochs for different splitting strategies.}
    \label{fig:Time}
\end{figure}

This benchmark allows us to compare the time consumption under different model splitting strategies. As mentioned earlier, we identified the slowest client as the system bottleneck and focus on its time consumption. Figure~\ref{fig:Time} shows the time expenditure of the slowest discriminator in the system over all epochs for different splitting strategies. We figure reports the time values averaged over all epochs along with the related standard deviations as error bars. We can observe that sorting the devices by the capacity first and assigning multiple portions to a single device, when possible, outperforms other splitting strategies. Moreover, we notice that selecting a device at random and assigning multiple parts to it yields the worst performance. This is due to the fact that devices with high available memory but low computing performance exists in our simulated environment. This is typical of old devices with high memory but lack of compatibility on latest instruction sets such as AVX, AVX2, or simply without GPU.

\fakeparagraph{Accuracy Benchmark} 
\begin{figure}[!t]
    \centering
    \includegraphics[width=\linewidth]{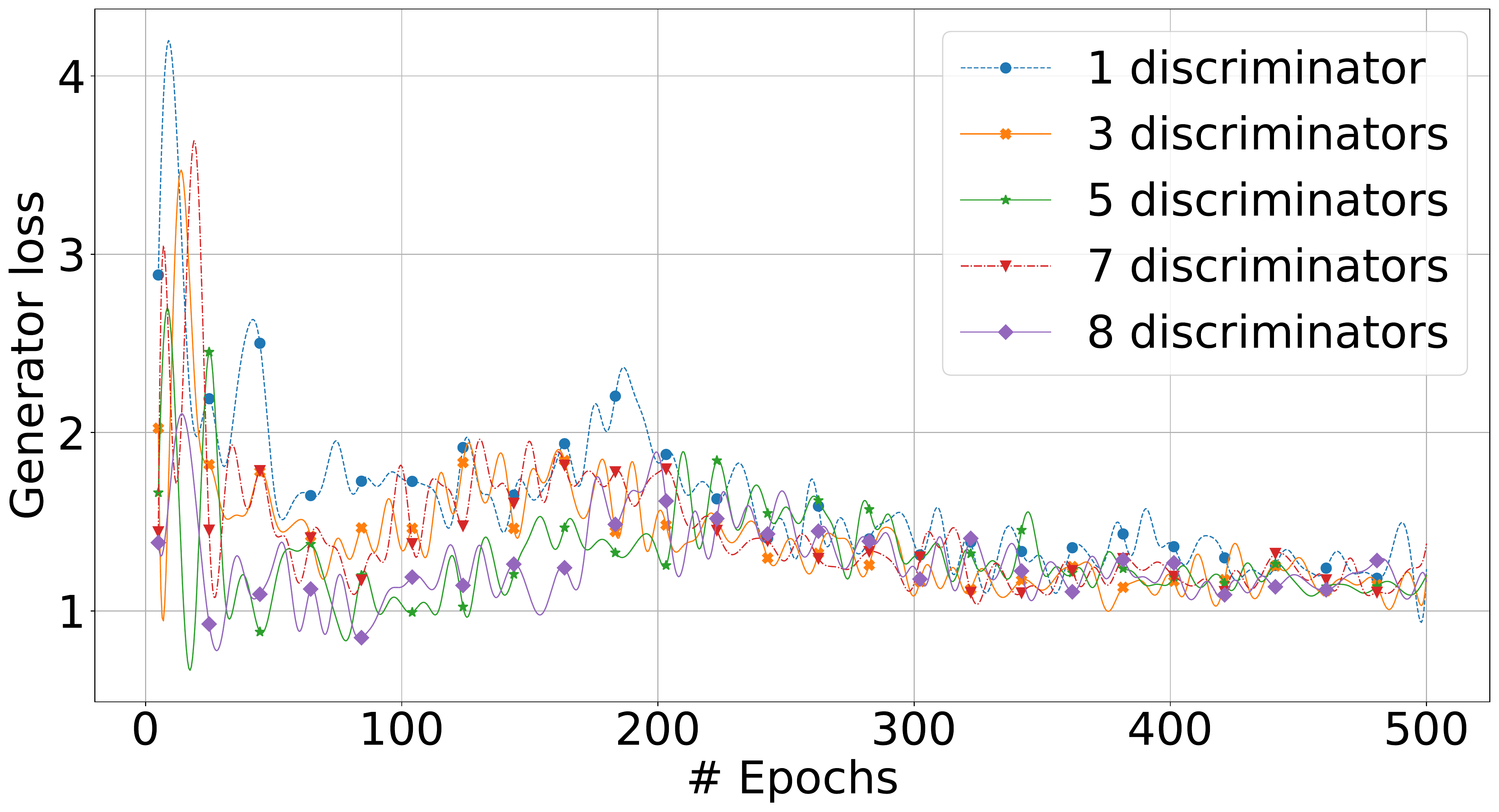}
    \caption{Generator loss for a varying number of discriminators as a function of the number of epochs.}
    \label{fig:Accuracy}
\end{figure}

\begin{figure}[!t]
    \centering
    \includegraphics[width=.9\linewidth]{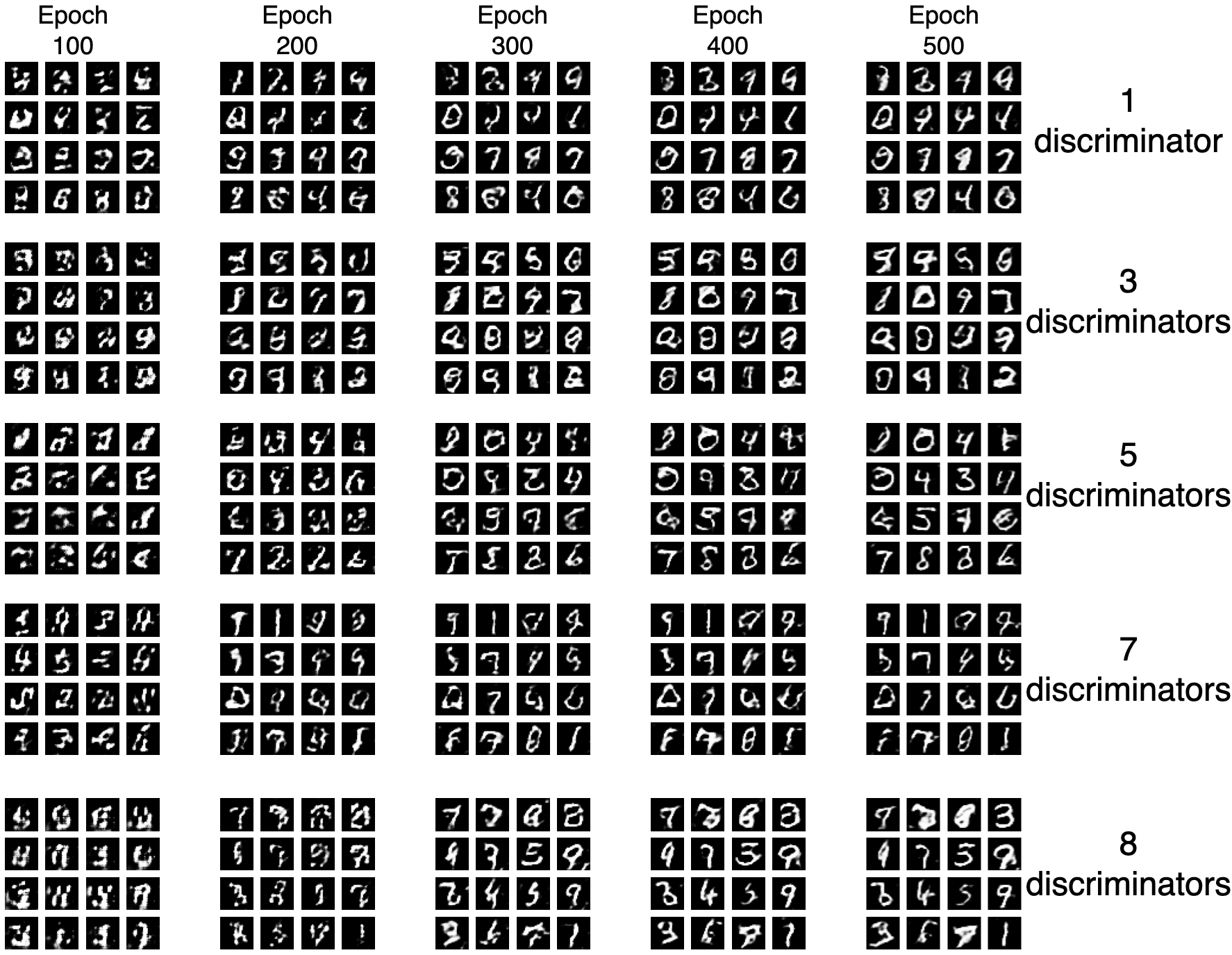}
    \caption{Image generated for a varying number of discriminators as a function of the number of epochs.}
    \label{fig:Generated_Images}
\end{figure}

Our accuracy benchmark allows us to evaluate whether the generator could converge under multiple discriminators instead of one. Figure~\ref{fig:Accuracy} shows the loss of the generator for a varying number of discriminators ($1$, $3$, $5$, $7$, and $8$) in the system. We notice that despite the fact that all the cases have concurrent fluctuations, a higher number of discriminators helps the generator to minimize its loss. Figure~\ref{fig:Generated_Images} depicts a set of images generated by the generator in a system with different number of discriminators as a function of the number of epochs. The results show that a higher number of discriminators does not necessarily yield to a faster convergence time; however, it does preserve the heterogeneity of the data distributions, which is a fundamental characteristic of FL.

\section{Conclusions}
\label{sec:conclusions}

FL and SL bring forth new opportunities to ensure data and model privacy. As such, it enables a whole multitude of applications that are built upon privacy-sensitive data. The proliferation of these applications is such that it often requires training of computationally-heavy models such as GANs. In this article, we envision a combination of both FL and SL that enables the training of GANs in resource-constrained devices. The proposed solution consists of collaboratively training the discriminator model of a GAN among different devices of a single client by fully leveraging their heterogeneous resources. In fact, we propose a heuristic device selection method that reflects their capabilities and helps us achieve better system performance. Specifically, we show that by increasing the number of discriminators in the system we achieve a better convergence of the model, and thus achieve a lower generator loss. Moreover, this method also decreases the training time.

Future research work will further improve our device selection method such that we 
\begin{enumerate*}[label=(\roman*)]
    \item minimize the communication overhead among devices, 
    \item consider the trade-off between memory size and computational performance (e.g., the bandwidth and the number of cores on GPU/CPU), 
    \item eliminate the slowest discriminator in the system, and more broadly 
    \item investigate the effect of data heterogeneity in the model convergence.
\end{enumerate*}


\begin{thebibliography}{21}


\ifx \showCODEN    \undefined \def \showCODEN     #1{\unskip}     \fi
\ifx \showDOI      \undefined \def \showDOI       #1{#1}\fi
\ifx \showISBNx    \undefined \def \showISBNx     #1{\unskip}     \fi
\ifx \showISBNxiii \undefined \def \showISBNxiii  #1{\unskip}     \fi
\ifx \showISSN     \undefined \def \showISSN      #1{\unskip}     \fi
\ifx \showLCCN     \undefined \def \showLCCN      #1{\unskip}     \fi
\ifx \shownote     \undefined \def \shownote      #1{#1}          \fi
\ifx \showarticletitle \undefined \def \showarticletitle #1{#1}   \fi
\ifx \showURL      \undefined \def \showURL       {\relax}        \fi
\providecommand\bibfield[2]{#2}
\providecommand\bibinfo[2]{#2}
\providecommand\natexlab[1]{#1}
\providecommand\showeprint[2][]{arXiv:#2}

\bibitem[Abedi and Khan(2020)]%
        {abedi2020fedsl}
\bibfield{author}{\bibinfo{person}{Ali Abedi} {and} \bibinfo{person}{Shehroz~S
  Khan}.} \bibinfo{year}{2020}\natexlab{}.
\newblock \showarticletitle{{FedSL: Federated Split Learning on Distributed
  Sequential Data in Recurrent Neural Networks}}.
\newblock \bibinfo{journal}{\emph{arXiv preprint arXiv:2011.03180}}
  (\bibinfo{year}{2020}).
\newblock


\bibitem[Cao et~al\mbox{.}(2018)]%
        {cao2018recent}
\bibfield{author}{\bibinfo{person}{Yang-Jie Cao}, \bibinfo{person}{Li-Li Jia},
  \bibinfo{person}{Yong-Xia Chen}, \bibinfo{person}{Nan Lin},
  \bibinfo{person}{Cong Yang}, \bibinfo{person}{Bo Zhang}, \bibinfo{person}{Zhi
  Liu}, \bibinfo{person}{Xue-Xiang Li}, {and} \bibinfo{person}{Hong-Hua Dai}.}
  \bibinfo{year}{2018}\natexlab{}.
\newblock \showarticletitle{Recent advances of generative adversarial networks
  in computer vision}.
\newblock \bibinfo{journal}{\emph{IEEE Access}}  \bibinfo{volume}{7}
  (\bibinfo{year}{2018}), \bibinfo{pages}{14985--15006}.
\newblock


\bibitem[Chang et~al\mbox{.}(2020)]%
        {chang2020multi}
\bibfield{author}{\bibinfo{person}{Qi Chang}, \bibinfo{person}{Zhennan Yan},
  \bibinfo{person}{Lohendran Baskaran}, \bibinfo{person}{Hui Qu},
  \bibinfo{person}{Yikai Zhang}, \bibinfo{person}{Tong Zhang},
  \bibinfo{person}{Shaoting Zhang}, {and} \bibinfo{person}{Dimitris~N
  Metaxas}.} \bibinfo{year}{2020}\natexlab{}.
\newblock \showarticletitle{{Multi-modal AsynDGAN: Learn From Distributed
  Medical Image Data without Sharing Private Information}}.
\newblock \bibinfo{journal}{\emph{arXiv preprint arXiv:2012.08604}}
  (\bibinfo{year}{2020}).
\newblock


\bibitem[Doolani et~al\mbox{.}(2020)]%
        {doolani2020review}
\bibfield{author}{\bibinfo{person}{Sanika Doolani}, \bibinfo{person}{Callen
  Wessels}, \bibinfo{person}{Varun Kanal}, \bibinfo{person}{Christos
  Sevastopoulos}, \bibinfo{person}{Ashish Jaiswal}, \bibinfo{person}{Harish
  Nambiappan}, {and} \bibinfo{person}{Fillia Makedon}.}
  \bibinfo{year}{2020}\natexlab{}.
\newblock \showarticletitle{A review of extended reality (xr) technologies for
  manufacturing training}.
\newblock \bibinfo{journal}{\emph{Technologies}} \bibinfo{volume}{8},
  \bibinfo{number}{4} (\bibinfo{year}{2020}), \bibinfo{pages}{77}.
\newblock


\bibitem[Gattullo et~al\mbox{.}(2019)]%
        {gattullo2019towards}
\bibfield{author}{\bibinfo{person}{Michele Gattullo},
  \bibinfo{person}{Giulia~Wally Scurati}, \bibinfo{person}{Michele Fiorentino},
  \bibinfo{person}{Antonio~Emmanuele Uva}, \bibinfo{person}{Francesco Ferrise},
  {and} \bibinfo{person}{Monica Bordegoni}.} \bibinfo{year}{2019}\natexlab{}.
\newblock \showarticletitle{Towards augmented reality manuals for industry 4.0:
  A methodology}.
\newblock \bibinfo{journal}{\emph{Robotics and Computer-Integrated
  Manufacturing}}  \bibinfo{volume}{56} (\bibinfo{year}{2019}),
  \bibinfo{pages}{276--286}.
\newblock


\bibitem[Goodfellow et~al\mbox{.}(2020)]%
        {goodfellow2020generative}
\bibfield{author}{\bibinfo{person}{Ian Goodfellow}, \bibinfo{person}{Jean
  Pouget-Abadie}, \bibinfo{person}{Mehdi Mirza}, \bibinfo{person}{Bing Xu},
  \bibinfo{person}{David Warde-Farley}, \bibinfo{person}{Sherjil Ozair},
  \bibinfo{person}{Aaron Courville}, {and} \bibinfo{person}{Yoshua Bengio}.}
  \bibinfo{year}{2020}\natexlab{}.
\newblock \showarticletitle{Generative adversarial networks}.
\newblock \bibinfo{journal}{\emph{Commun. ACM}} \bibinfo{volume}{63},
  \bibinfo{number}{11} (\bibinfo{year}{2020}), \bibinfo{pages}{139--144}.
\newblock


\bibitem[Hsu et~al\mbox{.}(2017)]%
        {hsu2017voice}
\bibfield{author}{\bibinfo{person}{Chin-Cheng Hsu}, \bibinfo{person}{Hsin-Te
  Hwang}, \bibinfo{person}{Yi-Chiao Wu}, \bibinfo{person}{Yu Tsao}, {and}
  \bibinfo{person}{Hsin-Min Wang}.} \bibinfo{year}{2017}\natexlab{}.
\newblock \showarticletitle{Voice conversion from unaligned corpora using
  variational autoencoding wasserstein generative adversarial networks}.
\newblock \bibinfo{journal}{\emph{arXiv preprint arXiv:1704.00849}}
  (\bibinfo{year}{2017}).
\newblock


\bibitem[Jabbar et~al\mbox{.}(2021)]%
        {jabbar2021survey}
\bibfield{author}{\bibinfo{person}{Abdul Jabbar}, \bibinfo{person}{Xi Li},
  {and} \bibinfo{person}{Bourahla Omar}.} \bibinfo{year}{2021}\natexlab{}.
\newblock \showarticletitle{A survey on generative adversarial networks:
  Variants, applications, and training}.
\newblock \bibinfo{journal}{\emph{ACM Computing Surveys (CSUR)}}
  \bibinfo{volume}{54}, \bibinfo{number}{8} (\bibinfo{year}{2021}),
  \bibinfo{pages}{1--49}.
\newblock


\bibitem[Jeon and Kim(2020)]%
        {jeon2020privacy}
\bibfield{author}{\bibinfo{person}{Joohyung Jeon} {and}
  \bibinfo{person}{Joongheon Kim}.} \bibinfo{year}{2020}\natexlab{}.
\newblock \showarticletitle{Privacy-Sensitive Parallel Split Learning}. In
  \bibinfo{booktitle}{\emph{2020 International Conference on Information
  Networking (ICOIN)}}. IEEE, \bibinfo{pages}{7--9}.
\newblock


\bibitem[Kairouz et~al\mbox{.}(2019)]%
        {kairouz2019advances}
\bibfield{author}{\bibinfo{person}{Peter Kairouz}, \bibinfo{person}{H~Brendan
  McMahan}, \bibinfo{person}{Brendan Avent}, \bibinfo{person}{Aur{\'e}lien
  Bellet}, \bibinfo{person}{Mehdi Bennis}, \bibinfo{person}{Arjun~Nitin
  Bhagoji}, \bibinfo{person}{Keith Bonawitz}, \bibinfo{person}{Zachary
  Charles}, \bibinfo{person}{Graham Cormode}, \bibinfo{person}{Rachel
  Cummings}, {et~al\mbox{.}}} \bibinfo{year}{2019}\natexlab{}.
\newblock \showarticletitle{Advances and open problems in federated learning}.
\newblock \bibinfo{journal}{\emph{arXiv preprint arXiv:1912.04977}}
  (\bibinfo{year}{2019}).
\newblock


\bibitem[Lin et~al\mbox{.}(2017)]%
        {lin2017adversarial}
\bibfield{author}{\bibinfo{person}{Kevin Lin}, \bibinfo{person}{Dianqi Li},
  \bibinfo{person}{Xiaodong He}, \bibinfo{person}{Zhengyou Zhang}, {and}
  \bibinfo{person}{Ming-Ting Sun}.} \bibinfo{year}{2017}\natexlab{}.
\newblock \showarticletitle{Adversarial ranking for language generation}.
\newblock \bibinfo{journal}{\emph{Advances in neural information processing
  systems}}  \bibinfo{volume}{30} (\bibinfo{year}{2017}).
\newblock


\bibitem[Liu et~al\mbox{.}(2019)]%
        {liu2019ppgan}
\bibfield{author}{\bibinfo{person}{Yi Liu}, \bibinfo{person}{Jialiang Peng},
  \bibinfo{person}{JQ James}, {and} \bibinfo{person}{Yi Wu}.}
  \bibinfo{year}{2019}\natexlab{}.
\newblock \showarticletitle{PPGAN: Privacy-preserving generative adversarial
  network}. In \bibinfo{booktitle}{\emph{2019 IEEE 25Th international
  conference on parallel and distributed systems (ICPADS)}}. IEEE,
  \bibinfo{pages}{985--989}.
\newblock


\bibitem[McMahan et~al\mbox{.}(2017)]%
        {mcmahan2017communication}
\bibfield{author}{\bibinfo{person}{Brendan McMahan}, \bibinfo{person}{Eider
  Moore}, \bibinfo{person}{Daniel Ramage}, \bibinfo{person}{Seth Hampson},
  {and} \bibinfo{person}{Blaise~Aguera y Arcas}.}
  \bibinfo{year}{2017}\natexlab{}.
\newblock \showarticletitle{Communication-efficient learning of deep networks
  from decentralized data}. In \bibinfo{booktitle}{\emph{Artificial
  intelligence and statistics}}. PMLR, \bibinfo{pages}{1273--1282}.
\newblock


\bibitem[Mogren(2016)]%
        {mogren2016c}
\bibfield{author}{\bibinfo{person}{Olof Mogren}.}
  \bibinfo{year}{2016}\natexlab{}.
\newblock \showarticletitle{{C-RNN-GAN: Continuous recurrent neural networks
  with adversarial training}}.
\newblock \bibinfo{journal}{\emph{arXiv preprint arXiv:1611.09904}}
  (\bibinfo{year}{2016}).
\newblock


\bibitem[Radford et~al\mbox{.}(2016)]%
        {DBLP:journals/corr/RadfordMC15}
\bibfield{author}{\bibinfo{person}{Alec Radford}, \bibinfo{person}{Luke Metz},
  {and} \bibinfo{person}{Soumith Chintala}.} \bibinfo{year}{2016}\natexlab{}.
\newblock \showarticletitle{Unsupervised Representation Learning with Deep
  Convolutional Generative Adversarial Networks}. In
  \bibinfo{booktitle}{\emph{4th International Conference on Learning
  Representations, {ICLR} 2016, San Juan, Puerto Rico, May 2-4, 2016,
  Conference Track Proceedings}}, \bibfield{editor}{\bibinfo{person}{Yoshua
  Bengio} {and} \bibinfo{person}{Yann LeCun}} (Eds.).
\newblock
\urldef\tempurl%
\url{http://arxiv.org/abs/1511.06434}
\showURL{%
\tempurl}


\bibitem[Rasouli et~al\mbox{.}(2020)]%
        {rasouli2020fedgan}
\bibfield{author}{\bibinfo{person}{Mohammad Rasouli}, \bibinfo{person}{Tao
  Sun}, {and} \bibinfo{person}{Ram Rajagopal}.}
  \bibinfo{year}{2020}\natexlab{}.
\newblock \showarticletitle{FedGAN: Federated generative adversarial networks
  for distributed data}.
\newblock \bibinfo{journal}{\emph{arXiv preprint arXiv:2006.07228}}
  (\bibinfo{year}{2020}).
\newblock


\bibitem[Singh and Chatterjee(2017)]%
        {singh2017cloud}
\bibfield{author}{\bibinfo{person}{Ashish Singh} {and} \bibinfo{person}{Kakali
  Chatterjee}.} \bibinfo{year}{2017}\natexlab{}.
\newblock \showarticletitle{Cloud security issues and challenges: A survey}.
\newblock \bibinfo{journal}{\emph{Journal of Network and Computer
  Applications}}  \bibinfo{volume}{79} (\bibinfo{year}{2017}),
  \bibinfo{pages}{88--115}.
\newblock


\bibitem[Thapa et~al\mbox{.}(2020)]%
        {thapa2020splitfed}
\bibfield{author}{\bibinfo{person}{Chandra Thapa}, \bibinfo{person}{Mahawaga
  Arachchige~Pathum Chamikara}, {and} \bibinfo{person}{Seyit Camtepe}.}
  \bibinfo{year}{2020}\natexlab{}.
\newblock \showarticletitle{Splitfed: When federated learning meets split
  learning}.
\newblock \bibinfo{journal}{\emph{arXiv preprint arXiv:2004.12088}}
  (\bibinfo{year}{2020}).
\newblock


\bibitem[Turina et~al\mbox{.}(2020)]%
        {turina2020combining}
\bibfield{author}{\bibinfo{person}{Valeria Turina}, \bibinfo{person}{Zongshun
  Zhang}, \bibinfo{person}{Flavio Esposito}, {and} \bibinfo{person}{Ibrahim
  Matta}.} \bibinfo{year}{2020}\natexlab{}.
\newblock \showarticletitle{Combining split and federated architectures for
  efficiency and privacy in deep learning}. In
  \bibinfo{booktitle}{\emph{Proceedings of the 16th International Conference on
  emerging Networking EXperiments and Technologies}}.
  \bibinfo{pages}{562--563}.
\newblock


\bibitem[Zhou et~al\mbox{.}(2020)]%
        {zhou2020edge}
\bibfield{author}{\bibinfo{person}{Pengyuan Zhou}, \bibinfo{person}{Tristan
  Braud}, \bibinfo{person}{Aleksandr Zavodovski}, \bibinfo{person}{Zhi Liu},
  \bibinfo{person}{Xianfu Chen}, \bibinfo{person}{Pan Hui}, {and}
  \bibinfo{person}{Jussi Kangasharju}.} \bibinfo{year}{2020}\natexlab{}.
\newblock \showarticletitle{Edge-facilitated augmented vision in
  vehicle-to-everything networks}.
\newblock \bibinfo{journal}{\emph{IEEE Transactions on Vehicular Technology}}
  \bibinfo{volume}{69}, \bibinfo{number}{10} (\bibinfo{year}{2020}),
  \bibinfo{pages}{12187--12201}.
\newblock


\bibitem[Zhou et~al\mbox{.}(2022)]%
        {leftout2022}
\bibfield{author}{\bibinfo{person}{Pengyuan Zhou}, \bibinfo{person}{Hengwei
  Xu}, \bibinfo{person}{Likhang Lee}, \bibinfo{person}{Pei Fang}, {and}
  \bibinfo{person}{Pan Hui}.} \bibinfo{year}{2022}\natexlab{}.
\newblock \showarticletitle{Are You Left Out? An Efficient and Fair Federated
  Learning for Personalized Profiles on Wearable Devices of Inferior Networking
  Conditions}.
\newblock \bibinfo{journal}{\emph{Proceedings of the ACM on Interactive,
  Mobile, Wearable and Ubiquitous Technologies (IMWUT/UbiComp)}}
  \bibinfo{volume}{6}, \bibinfo{number}{3} (\bibinfo{year}{2022}).
\newblock


\end{thebibliography}
\end{document}